\documentclass{vgtc}                          

\graphicspath{{figures/}{pictures/}{images/}{./}} 

\usepackage{times}                     

\usepackage{mathptmx}                  

\usepackage{svg}

\usepackage{caption}

\definecolor{GoogleColor}{HTML}{0b57d0}

\onlineid{1342}

\vgtccategory{VIS 2025 VISxGenAI}

\vgtcinsertpkg

\title{Benchmark It Yourself (BIY): Preparing a Dataset and Benchmarking AI Models for Scatterplot-Related Tasks}

\author{
    João Palmeiro \thanks{e-mail: joao.palmeiro@feedzai.com} %
    \and Diogo Duarte \thanks{e-mail: diogo.duarte@feedzai.com} %
    \and Rita Costa \thanks{e-mail: rita.costa@feedzai.com} %
    \and Pedro Bizarro \thanks{e-mail: pedro.bizarro@feedzai.com}
}
\affiliation{\scriptsize Feedzai}

\teaser{
  \centering
  \includegraphics[width=\linewidth, alt={Preview of the new scatterplot dataset and a list of the main features of the dataset and benchmark. The top section shows six of the scatterplots paired with their annotated versions. These versions are layered with cluster bounding boxes, cluster centers, and outliers. From left to right, the first scatterplot is composed of six clusters spread diagonally from the top-left to the bottom-right corner; the second is composed of four scattered, elongated clusters and background noise; the third has one cluster in the top-left corner and four outliers near the top-right and bottom-right corners; the fourth has three clusters that form a triangle; the fifth has a set of exponentially related points; the sixth has two spread out clusters, one on top and one on the bottom, and background noise. The bottom section, on the left, notes that the dataset is a new synthetic dataset, composed of over 18,000 annotated scatterplots and 17 chart designs. In the middle, it lists that the benchmark is composed of 1725 scatterplots, 10 models from OpenAI and Google, 5 tasks, and 3 prompting strategies. On the right, there are several examples of actual model responses next to the question: "What performance?". The examples are: {"clusters": [], "outliers": []}, (5), (38), (2), (100), {"cluster_centers": [[278, 747], [768, 250]]}, (0), (1), {"outliers": [[50, 50], [950, 50], [950, 950], [50, 950]]}, and {"clusters": [[66, 27, 438, 224], [466, 370, 927, 570]]}.}]{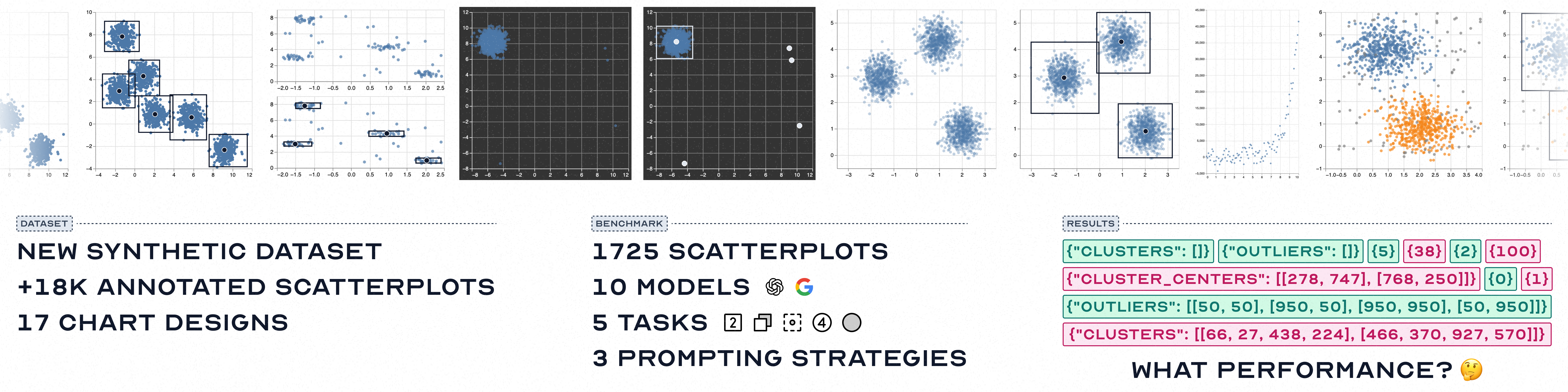}
  \caption{A high-level overview of the new dataset and benchmark for scatterplot-related tasks.}
  \label{fig:teaser}
}

\abstract{
AI models are increasingly used for data analysis and visualization, yet benchmarks rarely address scatterplot-specific tasks, limiting insight into performance. To address this gap for one of the most common chart types, we introduce a synthetic, annotated dataset of over 18,000 scatterplots from six data generators and 17 chart designs, and a benchmark based on it. We evaluate proprietary models from OpenAI and Google using N-shot prompting on five distinct tasks derived from annotations of cluster bounding boxes, their center coordinates, and outlier coordinates. OpenAI models and Gemini 2.5 Flash, especially when prompted with examples, are viable options for counting clusters and, in Flash's case, outliers (90\%+ Accuracy). However, the results for localization-related tasks are unsatisfactory: Precision and Recall are near or below 50\%, except for Flash in outlier identification (65.01\%). Furthermore, the impact of chart design on performance appears to be a secondary factor, but it is advisable to avoid scatterplots with wide aspect ratios (16:9 and 21:9) or those colored randomly. Supplementary materials are available at \url{https://github.com/feedzai/biy-paper}.
}


\begin{document}


\firstsection{Introduction}
\label{sec:intro}

\maketitle

Large Language Models (LLMs), particularly multimodal models, are among today's key digital technologies. From promising academic results~\cite{chart-citor,plot-gen} to new products and services~\cite{viz-gpt,manus-gallery,manus-paper}, these AI models have demonstrated wide-ranging applicability, including for data analysis and visualization.

Nevertheless, when implementing a feature or developing new ideas, the ever-increasing number of options (such as models, prompting strategies, and multi-agent architectures~\cite{bench-multi-agent-archs}) can be a problem. Experimenting with different approaches can be costly and time-consuming, and it ultimately takes expertise to make a sound decision. One could simply resort to a frontier model or adapt an existing multi-agent system; however, performance remains uncertain.

For data analysis and visualization, several benchmarks offer guidance~\cite{lmarena,vlm-eval-kit,dataset-survey,live-bench}. Yet, when it comes to scatterplots, they lack representativeness and are not targeted at scatterplot-related tasks such as clustering and outlier detection~\cite{dataset-survey,figureqa,chartqa-pro,chartqa,chart-insights,mmmu,mmmu-pro} — a significant gap, as scatterplots are one of the most commonly used chart types~\cite{beagle,soti-2024,scatterplot-history} and a standard way to visualize two numerical variables to extract insights from their patterns. Moreover, existing benchmarks can be manipulated~\cite{meta-gaming,leaderboard-illusion} or require nuanced interpretation~\cite{artificial-analysis-providers}, rendering existing numbers more uncertain. Furthermore, models with vision capabilities have been shown to be brittle, highlighting their weaknesses~\cite{do-you-see-me,perils-misleading-vlms,unraveling-the-truth,vlms-are-blind,alt-gosling,vlms-are-biased,llm-bar-chart-aligned}.

In this work, we explore the application of AI models to a set of typical visual scatterplot tasks~\cite{scatterplot-tasks} using a new synthetic, annotated dataset, contributing to the common understanding of the actual capabilities of current models. When there is no access to the raw data or when datasets are large and extensively plotted — especially on dynamic platforms without relevant a priori context for the data — relying on chart images may be the only viable approach for extracting insights. These insights can then be used to generate descriptions for the charts or to complement multi-agent systems for data analysis and reporting. Our \textbf{main contributions} are: 1) a synthetic, annotated dataset (and its generation pipeline) for scatterplot-related tasks (\ref{sec:bench-dataset}); 2) a comprehensive evaluation of the performance of ten proprietary models on said tasks (\ref{sec:bench-bench}); and 3) a list of considerations when designing charts and providing them as input to AI models (\ref{sec:considerations}).

\section{Related Work}
\label{sec:lit}

\subsection{Datasets}
\label{sec:datasets}

Multiple datasets are available for training and testing new models or for evaluating existing solutions. Chart understanding datasets are multimodal (each instance contains a chart image) and can be divided into categories according to task type. Three of the main ones are chart question answering (QA)~\cite{chart-check,mapqa,leaf-qa,pixmo-docs,better-chartqa,chartqa-x,opencqa,info-chartqa,sci-cqa,poly-chartqa,chart-bench,chartqa-pot,multi-chartqa}, chart captioning~\cite{hci-alt-texts,sci-cap,linecap,alt-chart,chart-to-text,owid,genchar,academia-chart}, and chart extraction~\cite{chart-info,orion-bench}.

PlotQA~\cite{plotqa}, alongside FigureQA~\cite{figureqa} and DVQA~\cite{dvqa}, is one of the first examples of chart QA datasets, consisting of a large set of annotations and question-answer pairs. VisText~\cite{vistext}, Alt4Blind~\cite{alt4blind}, and ChartSumm~\cite{chart-summ} are also rich in annotations but focus on chart descriptions or summaries. For chart extraction, ChartX~\cite{chartx} and SynthChartNet~\cite{synth-chart-net} map chart images to their corresponding tabular data (CSV) and a structured data format (OTSL~\cite{otsl}), respectively. There are also compilations, such as ChartSFT~\cite{chart-sft}, DataVizQA~\cite{datavizqa}, CHOCOLATE~\cite{chocolate}, and those assembled to pre- or post-train~\cite{instruction-tuning} models like Granite Vision~\cite{granite}, Llama Nemotron Nano VL~\cite{llama-nemotron}, and Seed1.5-VL~\cite{seed}. Lastly, datasets can be hybrid, combining chart understanding and chart code generation or chart extraction, for example~\cite{nova-chart,chart-galaxy,ref-chartqa,omni-chart,meta-chart,cosyn-400k}, or tasks across different domains like EMMA~\cite{emma}, MMMU~\cite{mmmu,mmmu-pro}, and KITAB-Bench~\cite{kitab-bench}.


Our dataset is a hybrid of chart QA and extraction. It stands out from existing datasets by including complementary, scatterplot-oriented annotations to evaluate the clustering and outlier detection performance of AI models (or to train them), as well as by the ease, accuracy, and negligible cost of extending it.

\subsection{Benchmarks and Experiments}
\label{sec:benchs}

Datasets like ChartQA~\cite{chartqa}, ChartQAPro~\cite{chartqa-pro}, CharXiv~\cite{charxiv}, ChartMuseum~\cite{chart-museum}, MMMU~\cite{mmmu}, and MMMU-Pro~\cite{mmmu-pro} are, in practice, benchmarks that are reported on when new models are released~\cite{qwen-vl-report,gemini-report,ovis-report,gpt-5,mimo-report,glm-report}. Each has an expected prompt structure (though the final prompt can be adjusted depending on the model~\cite{xiaomi-lmms-eval,mistral-evals,lmms-eval}) and at least one performance metric.

Beyond that, AI models have been evaluated from various perspectives, adding to the benchmark results. These experiments are as comprehensive as one can imagine, including visualization literacy testing~\cite{viz-test-gpt4,viz-test-do-llms,viz-help-ai,viz-test-comparative,viz-test-eurovis}, predicting the experiential impact of charts~\cite{chart-to-experience}, evaluating the (positive) contribution of including chart images for data analysis~\cite{viz-help-ai}, detecting misleading visualizations~\cite{how-good-detect-misleading-viz}, and even evaluating the performance of LLMs on various tasks, taking SVG charts as input instead of raster images~\cite{llms-svg}.

In any case, none of these benchmarks or experiments cover tasks related to multimodal (chart image + text) evaluation of scatterplots. The exceptions are PUB~\cite{pub} and ChartInsights~\cite{chart-insights}, the works closest to ours, which cover tasks for various chart types, including scatterplots. We chose to focus on scatterplots, tackling undocumented tasks such as outlier detection. We employ three different prompting strategies, a new set of models and metrics, and a distinct methodology for assembling the charts under evaluation. The scatterplots feature more complex patterns than the other works, resulting in a more diverse collection. Two other works~\cite{point-detect-count,bridging-vlm-evaluation} are also worth mentioning, given the similar experimental setup, although they address different domains.

\section{Benchmarking Scatterplot-Related Tasks}
\label{sec:bench}

In this section, we first introduce our dataset and the benchmark. Following that, we present the results and main considerations for future applications.

\subsection{Dataset}
\label{sec:bench-dataset}

Our synthetic dataset consists of 371 data samples, totaling \textbf{18,921 distinct scatterplots}. The data samples are obtained from six generators implemented in Python using NumPy~\cite{numpy}, pandas~\cite{pandas}, PyOD~\cite{pyod}, scikit-learn~\cite{scikit-learn}, SciPy~\cite{scipy}, and Shapely~\cite{shapely}. These data generators are designed to output Gaussian blobs with and without background noise (round clusters; 60 data samples each), Gaussian blobs with outliers (54), random patterns (scattered points with no clear pattern; 8), relationships (linear, exponential, and quadratic; 9), and, finally, geometric shape blobs (6 shapes; 180 in total). Each one is parameterized differently, with data samples containing between zero and six clusters. When outliers are injected, the contamination level varies between 0.001 and 0.01. We chose to inject a relatively small number of outliers, keeping them well-distanced from the clusters, in order to evaluate the detection of points that are clearly anomalous and relevant to report.

The scatterplot images are generated via Vega-Lite~\cite{vega-lite} and 17 different specifications~\cite{limitations-image-transforms,chart-text,chaos,unraveling-the-truth,chart-insights}. Each corresponds to a chart design created from the default styling, including itself (\textit{default}) \raisebox{-0.4ex}{\includesvg[height=2ex]{figures/scatterplot_default.svg}} and its dark-themed counterpart \raisebox{-0.4ex}{\includesvg[height=2ex]{figures/scatterplot_dark.svg}} (2):

\vspace{-\topsep}
\begin{itemize}
    \setlength{\parskip}{0pt}
    \item Aspect ratios (5): 3:4, 4:3, 9:16, 16:9, and 21:9.
    \item Colors (2): colored clusters and randomly colored points.
    \item Opacity (2): full opacity and half the default opacity (0.35 instead of 0.7).
    \item Chart elements (2): points only and Y-axis only.
    \item Point shapes (2): random shapes and squared points.
    \item Point size (2): half the default size (15 instead of 30) and twice the default size (60).
\end{itemize}
\vspace{-\topsep}

The pipeline to combine the data samples and the Vega-Lite specifications is built on top of FastHTML~\cite{fast-html}, Playwright~\cite{playwright}, and the Vega View API, generating an image and a set of annotations for each scatterplot. By creating a web application using FastHTML to render each scatterplot, launched in a separate thread, Playwright automatically interacts with this application to extract the scatterplot images and their corresponding annotations—all orchestrated from a Python script. The Vega View API is leveraged to transform Cartesian coordinates into screen coordinates (pixels) for each of the annotation types: cluster horizontal bounding boxes~\cite{earth-gpt}, cluster center coordinates, and outlier coordinates. Each image is saved in three sizes derived from Vega-Lite's standard dimensions: for a square image, approximately 150px, 300px (default), and 600px (the reference size used in the benchmark~\cite{vit-blog,paligemma,resolution-curse}).

Finally, the scatterplot images are optimized with Oxipng~\cite{oxipng} to reduce the total file size and remove unnecessary metadata.

\subsection{Benchmark}
\label{sec:bench-bench}

For the benchmark, the dataset was randomly sampled to focus on the stipulated analyses and balance costs. We tested \textbf{1,725 scatterplots}, featuring 15 chart designs (all except the point shape ones) for 115 data samples: 25 Gaussian blobs, 25 Gaussian blobs with background noise, 50 Gaussian blobs with outliers, 7 random patterns, and 8 relationships.

There were \textbf{five scatterplot-related tasks} under evaluation: cluster counting, cluster detection (via bounding boxes), cluster identification (from their respective centers), outlier counting, and outlier identification (from the points' coordinates). Each one was defined by a prompt structured with an instruction and a response format~\cite{vlms-are-blind,vlms-are-biased}:

\vspace{-\topsep}
\begin{itemize}
    \setlength{\parskip}{0pt}
    \item \raisebox{-0.25ex}{\includesvg[height=2ex]{figures/cluster_counting.svg}} \textbf{Cluster counting}: \textit{How many clusters are there in the scatterplot? Answer with a number in curly brackets, e.g., \{4\}.}
    \item \raisebox{-0.25ex}{\includesvg[height=2ex]{figures/cluster_detection.svg}} \textbf{Cluster detection}: \textit{Detect all clusters in the scatterplot. For each detected cluster, provide its bounding box in normalized coordinates ~\cite{gemini-bbox,computer-vision-tasks} (x1, y1, x2, y2), where (0, 0) is the top-left corner and (1000, 1000) is the bottom-right corner of the image. Answer with a JSON object, e.g., \{"clusters": [[200, 30, 300, 50]]\}.}
    \item \raisebox{-0.25ex}{\includesvg[height=2ex]{figures/cluster_identification.svg}} \textbf{Cluster identification}: \textit{Identify each cluster in the scatterplot. For each, provide its center point in normalized coordinates (x, y), where (0, 0) is the top-left corner and (1000, 1000) is the bottom-right corner of the image. Answer with a JSON object, e.g., \{"cluster\_centers": [[250, 40], [700, 500]]\}.}
    \item \raisebox{-0.25ex}{\includesvg[height=2ex]{figures/outlier_counting.svg}} \textbf{Outlier counting}: \textit{How many outliers are there in the scatterplot? Answer with a number in curly brackets, e.g., \{3\}.}
    \item \raisebox{-0.25ex}{\includesvg[height=2ex]{figures/outlier_identification.svg}} \textbf{Outlier identification}: \textit{Identify each outlier in the scatterplot. For each, provide its location in normalized coordinates (x, y), where (0, 0) is the top-left corner and (1000, 1000) is the bottom-right corner of the image. Answer with a JSON object, e.g., \{"outliers": [[150, 900], [820, 150]]\}.}
\end{itemize}
\vspace{-\topsep}

In addition to the \textbf{zero-shot prompts} listed above, two other prompting strategies were used: \textbf{one-shot} and \textbf{few-shot} in turn-based conversation format~\cite{multi-turn,n-shot-prompting}. To this end, six examples were randomly selected from the complete dataset, before it was sampled, to represent each of the benchmarked data generators (plus an extra one so that the few-shot prompts would include two examples with outliers). We chose to include these two prompting strategies because they are easy to prepare (assuming there is labeled data), they do not require a higher number of inferences as prompt chaining strategies do, and they are expected to yield positive results~\cite{simon-blog}.

The \textbf{models} tested are largely from OpenAI \raisebox{-0.25ex}{\includesvg[height=2ex]{figures/OpenAI.svg}}, as listed below. Their extensive adoption in various contexts~\cite{state-web-dev-ai,llm-design-study,bvi-genai-tools,state-ai} motivated us to focus on this provider. Nevertheless, we also tested two of Google's \raisebox{-0.25ex}{\includesvg[height=2ex]{figures/Google.svg}} models to begin comparing results across providers, especially for low-cost models.

\vspace{-\topsep}
\begin{itemize}
    \setlength{\parskip}{0pt}
    \item \raisebox{-0.5ex}{\includesvg[height=2ex]{figures/OpenAI.svg}} \textit{gpt-4.1-2025-04-14} (GPT-4.1), \textit{gpt-4.1-mini-2025-04-14} (GPT-4.1 mini), \textit{gpt-4.1-nano-2025-04-14} (GPT-4.1 nano), \textit{gpt-4o-2024-08-06} (GPT-4o), and \textit{gpt-4o-mini-2024-07-18} (GPT-4o mini) with the \textit{temperature} set to 0.
    \item \raisebox{-0.5ex}{\includesvg[height=2ex]{figures/OpenAI.svg}} \textit{o3-2025-04-16} (o3) and \textit{o4-mini-2025-04-16} (o4-mini) with the \textit{temperature} set to 1 and \textit{reasoning\_effort} to \textit{medium}.
    \item \raisebox{-0.5ex}{\includesvg[height=2ex]{figures/Google.svg}} \textit{gemini-2.5-flash} (\textcolor{GoogleColor}{Flash}) and \textit{gemini-2.5-flash-lite} (\textcolor{GoogleColor}{Flash-Lite}) with the \textit{temperature} set to 0 and no thinking\footnote{We use "thinking" and "reasoning" interchangeably.} budget.
    \item \raisebox{-0.5ex}{\includesvg[height=2ex]{figures/Google.svg}} \textit{gemini-2.5-flash-lite} (\textcolor{GoogleColor}{Flash-Lite (Thinking)}) with the \textit{temperature} set to 0 and thinking budget to 8192 tokens~\cite{openai-gemini,nu-markdown}.
\end{itemize}
\vspace{-\topsep}

All OpenAI models were used with the image input detail level set to \textit{high}, whereas Google's equivalent was set to \textit{MEDIA\_RESOLUTION\_MEDIUM}\footnote{\href{https://github.com/googleapis/python-genai/issues/1198}{github.com/googleapis/python-genai/issues/1198}}.

\subsection{Results}
\label{sec:results}

The benchmark was run between July 29th and August 5th, 2025, using the Batch APIs from OpenAI and Vertex AI (Google)~\cite{daily-bench}. The total cost was approximately \$666~\cite{computer-vision-tasks} for 258,750 requests ($115\; data\;samples \times 15\;chart\;designs \times 10\;models \times 5\;tasks \times 3\;prompting\;strategies$).

The raw responses were processed to extract the counts and denormalize the bounding boxes and points, according to the prompts. Between 70.2\% and 99.95\% of OpenAI's responses were considered valid per model, whereas for Google, the range was between 76.52\%\footnote{\href{https://github.com/googleapis/python-genai/issues/782}{github.com/googleapis/python-genai/issues/782}} and 91.91\%. When breaking down the responses by chart design, the values are between 89.17\% and 92.66\%.

\subsubsection{Counting}
\label{sec:results-counting}

\begin{figure}[h]
    \captionsetup{belowskip=-15pt}
    \centering
    \includegraphics[width=\linewidth, alt={This is a grouped vertical bar chart. It's title is performance for the cluster counting task. The y-axis legend is Accuracy. The x-axis legend is model. The chart is made up by 10 groups of bars: GPT-4.1, GPT-4.1 mini, GPT-4.1 nano, GPT-4o, GPT-4o mini, o3, o4-mini, Flash, Flash-Lite, Flash-Lite (Thinking). Each group contains 3 bars: zero-shot prompt, one-shot, few-shot, which will be presented in that order. The first group of bars is GPT-4.1 and has values 65.28\%, 85.57\%, 95.3\%. The second group of bars is GPT-4.1 mini and has values 57.16\%, 72.23\%, 95.71\%. The third group of bars is GPT-4.1 nano and has values 57.0\%, 51.23\%, 56.58\%. The fourth group of bars is GPT-4o and has values 44.58\%, 75.52\%, 90.2\%. The fifth group of bars is GPT-4o mini and has values 44.81\%, 42.2\%, 52.99\%. The sixth group of bars is o3 and has values 74.49\%, 83.25\%, 97.16\%. The seventh group of bars is o4-mini and has values 54.43\%, 62.38\%, 94.26\%. The eighth group of bars is Flash and has values 46.55\%, 56.23\%, 95.65\%. The ninth group of bars is Flash-Lite and has values 41.74\%, 39.98\%, 65.57\%. The tenth group of bars is Flash-Lite (Thinking) and has values 51.21\%, 42.74\%, 84.4\%.}]{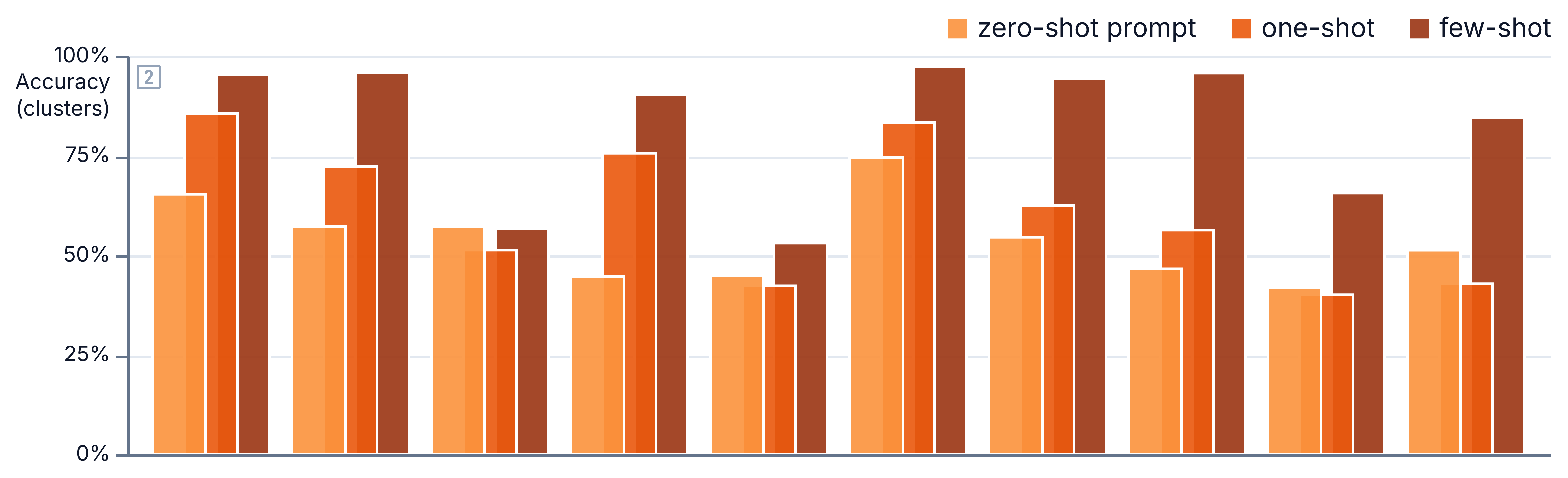}
    \includegraphics[width=\linewidth, alt={This is a grouped vertical bar chart. It's title is performance for the outlier counting task. The y-axis legend is Accuracy. The x-axis legend is model. The chart is made up by 10 groups of bars: GPT-4.1, GPT-4.1 mini, GPT-4.1 nano, GPT-4o, GPT-4o mini, o3, o4-mini, Flash, Flash-Lite, Flash-Lite (Thinking). Each group contains 3 bars: zero-shot prompt, one-shot, few-shot, which will be presented in that order. The first group of bars is GPT-4.1 and has values 52.93\%, 60.64\%, 78.43\%. The second group of bars is GPT-4.1 mini and has values 52.41\%, 72.7\%, 76.46\%. The third group of bars is GPT-4.1 nano and has values 32.93\%, 63.07\%, 67.07\%. The fourth group of bars is GPT-4o and has values 50.99\%, 75.94\%, 82.38\%. The fifth group of bars is GPT-4o mini and has values 61.62\%, 67.83\%, 63.01\%. The sixth group of bars is o3 and has values 49.45\%, 60.93\%, 56.64\%. The seventh group of bars is o4-mini and has values 39.73\%, 57.68\%, 57.91\%. The eighth group of bars is Flash and has values 54.65\%, 86.67\%, 90.49\%. The ninth group of bars is Flash-Lite and has values 60.5\%, 53.28\%, 59.88\%. The tenth group of bars is Flash-Lite (Thinking) and has values 66.67\%, 71.9\%, 68.09\%.}]{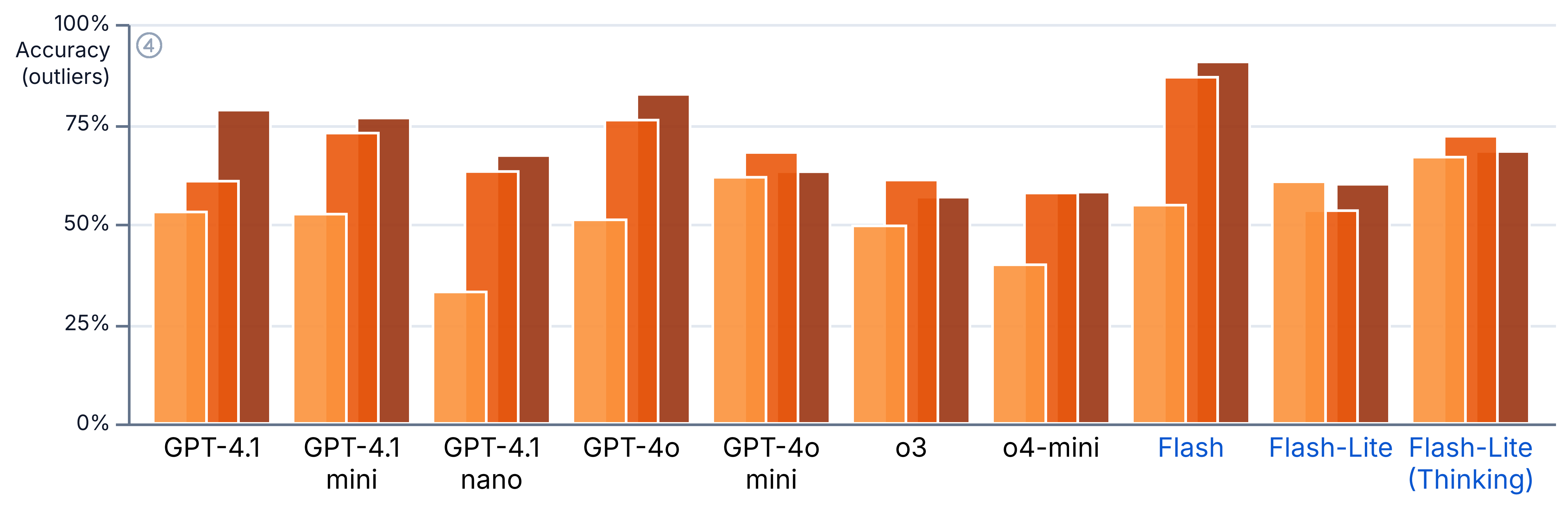}
    \caption{Accuracy for each model and prompting strategy. The results for cluster counting \raisebox{-0.4ex}{\includesvg[height=2ex]{figures/cluster_counting.svg}} (top) using few-shot prompting are particularly promising for several models. On the other hand, Flash excels at the outlier counting \raisebox{-0.4ex}{\includesvg[height=2ex]{figures/outlier_counting.svg}} task (bottom) when one-shot (86.67\%) and few-shot (90.49\%) prompted.}
    \label{fig:counting-accuracy}
\end{figure}

For the cluster and outlier counting tasks, we leverage two metrics: \textbf{Accuracy} (the proportion of correct counts) and ~\textbf{Mean Absolute Error (MAE)}~\cite{point-detect-count,mae}. The former allows us to estimate how good the models and prompting strategies are, while the latter to quantify the deviation in the predicted counts.

Figure~\ref{fig:counting-accuracy} shows the results for \textbf{Accuracy}. Regarding \textbf{MAE}, the best value for cluster counting is 0.03 (o3 when few-shot prompted), while the worst is 1.46 (Flash-Lite (Thinking) when one-shot prompted). However, although GPT-4o achieved an MAE of 0.29 for outlier counting when few-shot prompted, o4-mini achieves 18.54, which is much higher than expected given the relatively low number of outliers in the dataset.

On the other hand, the \textbf{consistency} in the predicted number of clusters \raisebox{-0.4ex}{\includesvg[height=2ex]{figures/cluster_counting.svg}} \raisebox{-0.4ex}{\includesvg[height=2ex]{figures/cluster_detection.svg}} \raisebox{-0.4ex}{\includesvg[height=2ex]{figures/cluster_identification.svg}} or outliers \raisebox{-0.4ex}{\includesvg[height=2ex]{figures/outlier_counting.svg}} \raisebox{-0.4ex}{\includesvg[height=2ex]{figures/outlier_identification.svg}} varied considerably across models and prompting strategies. Consistency implies that a given scatterplot is predicted with the same number of clusters/outliers, whether by counting them or by checking the number of predicted bounding boxes or coordinates. The results show that models like o3 and GPT-4.1, when few-shot prompted, predicted the same amount of clusters for over 90\% of scatterplots. Others, such as GPT-4.1 nano and mini (both zero-shot prompted), responded with the same number of clusters and outliers only 26.71\% and 24.39\% of the time, respectively. On average, only 61.4\% of all scatterplots were predicted with the same number of clusters (57.7\% for outliers).

The results exclusively for scatterplots with \textbf{no clustering patterns or outliers} also vary considerably. When the correct answer is simply \textit{0}, GPT-4.1, o3, and Flash, when few-shot prompted, achieve 100\% Accuracy in cluster counting. Flash, when one-shot and few-shot prompted, reaches 100\% for outlier counting as well. However, in cluster counting, Accuracy is below 1\% for Flash and Flash-Lite (one-shot), as well as GPT-4o (zero-shot). The lowest Accuracy in outlier counting is 50.89\% for the zero-shot-prompted o4-mini.

\subsubsection{Detection and Identification}
\label{sec:results-detection}

\begin{figure}[h]
    \captionsetup{belowskip=-15pt}
    \centering
    \includegraphics[width=\linewidth, alt={This is a grouped vertical bar chart. It's title is performance for the cluster detection task. The y-axis legend is Recall @ IoU75. The x-axis legend is model. The chart is made up by 10 groups of bars: GPT-4.1, GPT-4.1 mini, GPT-4.1 nano, GPT-4o, GPT-4o mini, o3, o4-mini, Flash, Flash-Lite, Flash-Lite (Thinking). Each group contains 3 bars: zero-shot prompt, one-shot, few-shot, which will be presented in that order. The first group of bars is GPT-4.1 and has values 5.69\%, 3.76\%, 18.88\%. The second group of bars is GPT-4.1 mini and has values 3.64\%, 10.23\%, 17.99\%. The third group of bars is GPT-4.1 nano and has values 1.52\%, 1.1\%, 10.54\%. The fourth group of bars is GPT-4o and has values 5.38\%, 8.08\%, 19.25\%. The fifth group of bars is GPT-4o mini and has values 0.35\%, 2.23\%, 10.45\%. The sixth group of bars is o3 and has values 7.53\%, 9.52\%, 19.91\%. The seventh group of bars is o4-mini and has values 8.28\%, 11.9\%, 18.07\%. The eighth group of bars is Flash and has values 9.37\%, 5.2\%, 18.65\%. The ninth group of bars is Flash-Lite and has values 6.11\%, 2.9\%, 18.96\%. The tenth group of bars is Flash-Lite (Thinking) and has values 11.03\%, 14.24\%, 16.77\%.}]{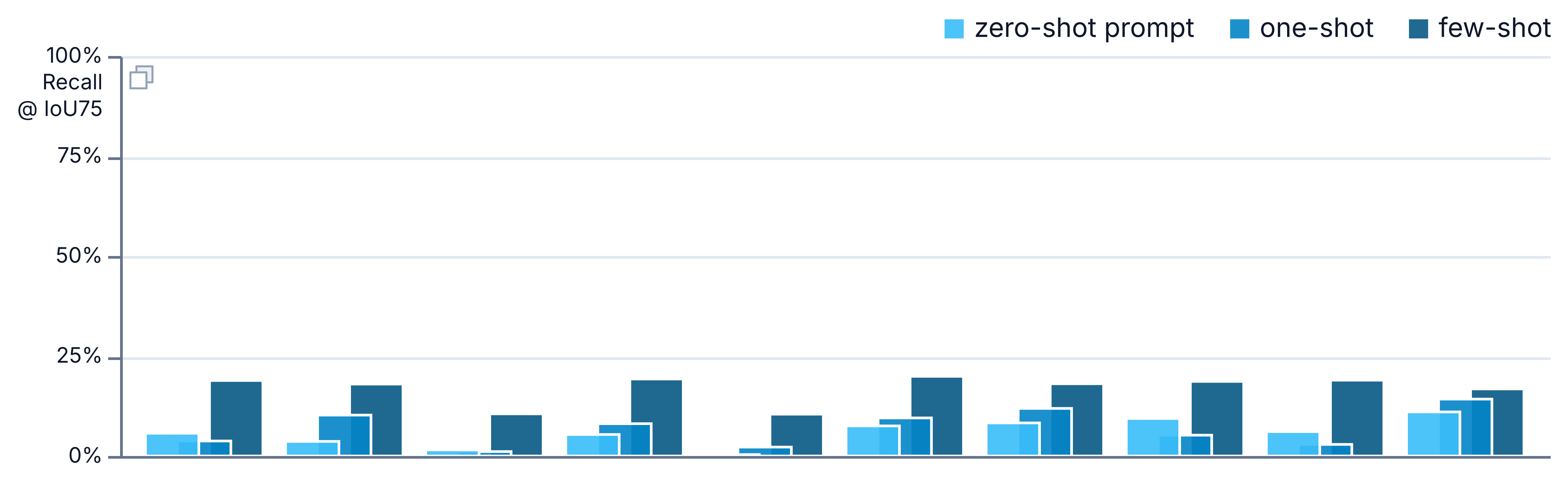}
    \includegraphics[width=0.9825\linewidth, alt={This is a grouped vertical bar chart. It's title is performance for the cluster identification task. The y-axis legend is Recall @ 10px. The x-axis legend is model. The chart is made up by 10 groups of bars: GPT-4.1, GPT-4.1 mini, GPT-4.1 nano, GPT-4o, GPT-4o mini, o3, o4-mini, Flash, Flash-Lite, Flash-Lite (Thinking). Each group contains 3 bars: zero-shot prompt, one-shot, few-shot, which will be presented in that order. The first group of bars is GPT-4.1 and has values 2.31\%, 5.52\%, 20.27\%. The second group of bars is GPT-4.1 mini and has values 1.89\%, 7.46\%, 20.2\%. The third group of bars is GPT-4.1 nano and has values 0.62\%, 2.85\%, 6.73\%. The fourth group of bars is GPT-4o and has values 1.64\%, 7.02\%, 18.63\%. The fifth group of bars is GPT-4o mini and has values 0.31\%, 2.27\%, 10.58\%. The sixth group of bars is o3 and has values 7.26\%, 6.06\%, 17.18\%. The seventh group of bars is o4-mini and has values 7.69\%, 12.84\%, 17.28\%. The eighth group of bars is Flash and has values 3.0\%, 2.78\%, 17.83\%. The ninth group of bars is Flash-Lite and has values 3.04\%, 3.03\%, 17.64\%. The tenth group of bars is Flash-Lite (Thinking) and has values 5.65\%, 7.57\%, 14.92\%.}]{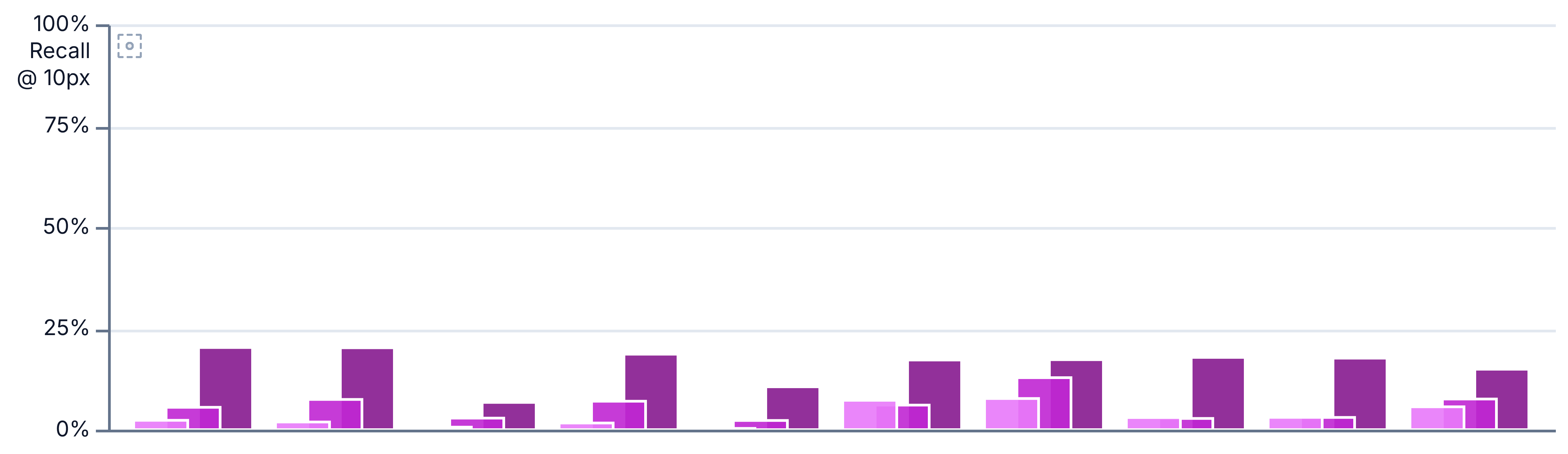}
    \includegraphics[width=0.9825\linewidth, alt={This is a grouped vertical bar chart. It's title is performance for the outlier identification task. The y-axis legend is Recall @ 10px. The x-axis legend is model. The chart is made up by 10 groups of bars: GPT-4.1, GPT-4.1 mini, GPT-4.1 nano, GPT-4o, GPT-4o mini, o3, o4-mini, Flash, Flash-Lite, Flash-Lite (Thinking). Each group contains 3 bars: zero-shot prompt, one-shot, few-shot, which will be presented in that order. The first group of bars is GPT-4.1 and has values 37.71\%, 17.35\%, 47.56\%. The second group of bars is GPT-4.1 mini and has values 1.31\%, 16.43\%, 40.22\%. The third group of bars is GPT-4.1 nano and has values 4.08\%, 24.86\%, 42.16\%. The fourth group of bars is GPT-4o and has values 9.89\%, 50.06\%, 56.83\%. The fifth group of bars is GPT-4o mini and has values 17.85\%, 37.08\%, 40.88\%. The sixth group of bars is o3 and has values 24.77\%, 37.45\%, 38.13\%. The seventh group of bars is o4-mini and has values 7.88\%, 15.13\%, 36.64\%. The eighth group of bars is Flash and has values 30.05\%, 45.15\%, 65.01\%. The ninth group of bars is Flash-Lite and has values 32.19\%, 37.34\%, 38.78\%. The tenth group of bars is Flash-Lite (Thinking) and has values 17.69\%, 33.16\%, 26.75\%.}]{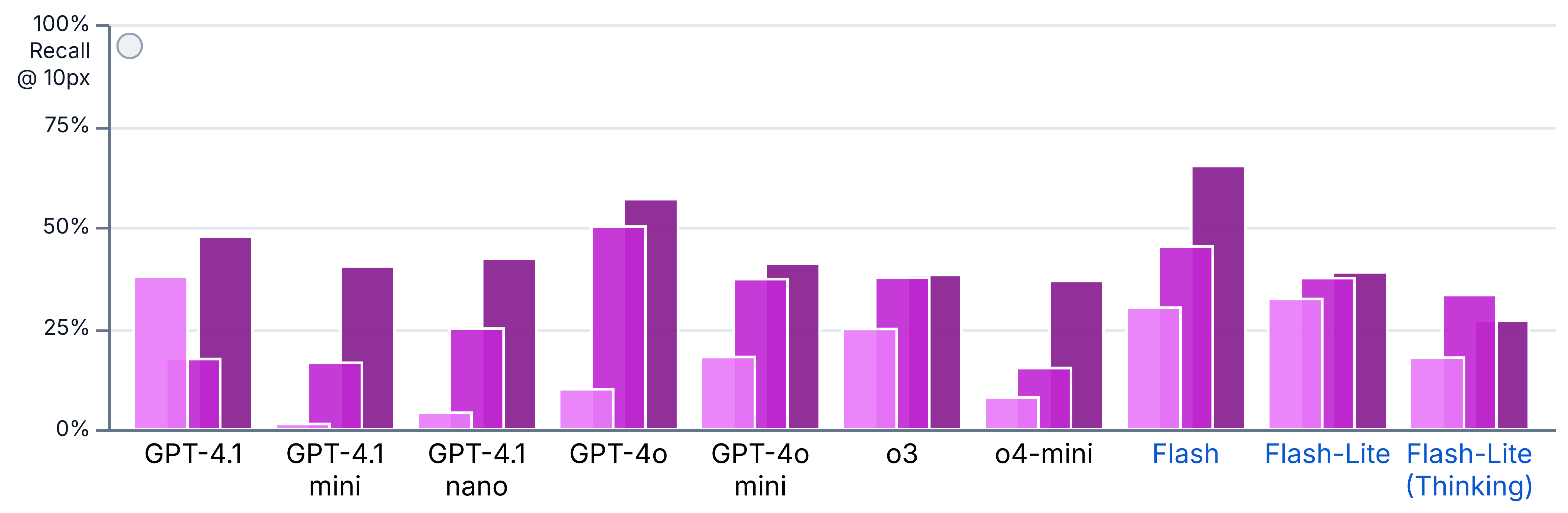}
    \caption{Recall for each model and prompting strategy. None surpass 25\% Recall in the cluster detection \raisebox{-0.4ex}{\includesvg[height=2ex]{figures/cluster_detection.svg}} (top) and identification \raisebox{-0.4ex}{\includesvg[height=2ex]{figures/cluster_identification.svg}} (middle) tasks. Recall is also low for the outlier identification \raisebox{-0.4ex}{\includesvg[height=2ex]{figures/outlier_identification.svg}} task (bottom), although Flash, when few-shot prompted, seems promising (65.01\%). Lowering the IoU to 0.5 considerably changes the results for some models, but Recall remains low (o3, when few-shot prompted, reaches a Recall of 50.26\%). Doubling the distance threshold to 20px does not change the conclusions.}
    \label{fig:recall-bboxes-points}
\end{figure}

For detection and identification tasks involving bounding boxes and point coordinates, we rely on \textbf{Precision} and \textbf{Recall}~\cite{visual-text-grounding,precision-recall}. More specifically, after obtaining the Precision and Recall values for each individual scatterplot, we average and report them considering the relevant groupings~\cite{deep-forest}. These two metrics allow us to verify if only relevant objects are being predicted, as well as if all relevant objects are indeed predicted. Plus, these metrics do not rely on models reporting confidence scores~\cite{deep-forest,earth-gpt}.

For cluster detection, we start by matching predictions and targets based on their Intersection over Union (IoU) values~\cite{lap,iou}. Then, we compute the confusion matrix. A True Positive corresponds to a predicted bounding box with an IoU of 0.75 or higher (\textbf{Precision and Recall @ IoU75}). Extra bounding boxes are considered False Positives, and if a target is not matched, we consider it a False Negative. This approach is also applied to identification tasks, where the threshold is a Euclidean distance of 10px~\cite{point-detect-count} (\textbf{Precision and Recall @ 10px}). We opted for strict thresholds to determine whether a given model and prompting strategy can be used when numerical precision is required.

Figure~\ref{fig:recall-bboxes-points} shows the results for \textbf{Recall} (the distribution of Precision values is similar across models and prompting strategies). The results for scatterplots with \textbf{no clusters or outliers} vary significantly again depending on the model and prompting strategy. When the correct answer is an empty list, GPT-4.1, GPT-4o, and Flash, when few-shot prompted, respond correctly between 98.22\% and 100\% of the time on the various detection and identification tasks. However, particularly in the cluster identification task, five of the seven OpenAI models, when zero-shot or one-shot prompted, score as low as 0\%.

\subsubsection{Chart Design Performance}
\label{sec:results-chart-design}

The motivation for benchmarking 15 chart designs stems from the need to evaluate the robustness of these models (three clusters should always be three clusters, regardless of the chart's aspect ratio or colors, for example) and to identify potential aesthetic choices that appear to negatively impact performance, which is often mitigable.

In the cluster counting task, the one with the best overall results, the best model when few-shot prompted (o3) achieves an Accuracy of 99.13\% on the \textit{default} chart design examples and on four others. However, for the \textit{points\_only} chart design, the Accuracy is only 90.43\% — a gap of 8.7 percentage points (pp). These differences vary across models, prompting strategies, and chart designs.


Therefore, we evaluated the \textbf{effect of chart design on cluster counting accuracy} using logistic regression with the \textit{default} chart design as the reference category. While the model's overall explanatory power~\cite{performance-package} was low ($pseudo\;R^2 = 0.0013$), a likelihood ratio test indicated that chart design significantly improved model fit compared to the null model ($p < 0.001$). Relative to the \textit{default} chart design (66.94\% accuracy), the \textit{half\_opacity} design significantly improved accuracy by 2.42 pp (69.36\%, $odds\;ratio\;(OR) = 1.12$, $p = 0.032$). Conversely, three chart designs significantly impaired performance: \textit{16\_9} (64.45\%, $OR = 0.9$, $p = 0.03$), \textit{21\_9} (62.34\%, $OR = 0.82$, $p < 0.001$), and \textit{random\_colors} (62.01\%, $OR = 0.81$, $p < 0.001$). These findings suggest that although chart design may explain a small proportion of the overall variability in accuracy, individual design choices can have nontrivial effects on performance.

\subsection{Considerations}
\label{sec:considerations}

Based on the results, the main considerations when combining scatterplot images and AI models are as follows:

\vspace{-\topsep}
\begin{enumerate}
    \setlength{\parskip}{0pt}
    \item \textbf{Prioritize few-shot prompting.} Few-shot prompting consistently outperformed zero-shot prompting across all models and tasks, with top models (Figure~\ref{fig:counting-accuracy}) achieving over 90\% Accuracy in counting tasks. Including examples is also useful for handling zero-answer scatterplots. Regardless, it can be worthwhile to determine the ideal number of examples for a given task to balance costs.
    \item \textbf{Avoid using OpenAI and low-cost Google models with similar prompting strategies for localization tasks involving scatterplots.} Results are significantly better for counting tasks than for precise detection/identification tasks (Figure~\ref{fig:recall-bboxes-points}), where their performance is unreliable, compromising their usefulness.
    \item \textbf{Invest in other components first, not chart design.} Chart design is fundamental for humans, but it's a secondary factor when fed into AI models. Even so, it can be beneficial to avoid chart designs with wide aspect ratios (16:9 and 21:9) or seemingly random colors. Adding opacity to points (for example, 0.35 in Vega-Lite and Vega-Altair~\cite{vega-altair}) can also be helpful.
\end{enumerate}
\vspace{-\topsep}

\section{Conclusion and Future Work}
\label{sec:future}

In this work, we address the lack of benchmarks for scatterplot-related tasks by introducing a novel synthetic dataset and a comprehensive benchmark. This is our first step toward better gauging the real capabilities of AI models for chart understanding and their applicability to real-world contexts. Consequently, we intend to further study the impact of chart design on performance, analyze the trade-off between cost and performance, and examine invalid responses.

Subsequently, we plan to include more data generators, chart designs, and internal data, as well as enhance the benchmark's representativeness. To this end, we intend to incorporate new prompting strategies, tasks (such as identifying relationships between variables and cluster shapes), and proprietary and open~\cite{open-models} models. In parallel, we also plan to create a lightweight version of the dataset and benchmark~\cite{emma,tiny-benchmarks} to monitor the performance of new models and prompting strategies at low cost.


Following this phase, we will proceed to fine-tune small, open models~\cite{vision-finetune} and further assess the trade-offs of using significantly less demanding models that can be run on our own infrastructure.

Finally, we will focus on chart captioning, leveraging the dataset to generate template-based descriptions for accessible scatterplots. Our goal is to evaluate the use of AI models to generate chart descriptions that can be used as \verb|alt| text for chart images (for instance, when the data is plotted using the Canvas API~\cite{svg-vs-canvas} due to the dataset's size)~\cite{alt-text-data-plots,telephone,mit-framework,matplotalt,bar-chart-templates,do-not-harm-guide,alt-gosling}. Thus, extending our work to other chart types will only come later.

\acknowledgments{
First, we would like to thank Iker Perez Lopez, Jean V. Alves, Javier Liébana, and Pedro Silva for all their help. We are also grateful to the reviewers for their comments.

We acknowledge the use of Kagi Translate for text translation and proofreading, as well as Cursor for development. The font used in the teaser is Optician Sans. The task icons are adapted from Phosphor.
}


\bibliographystyle{abbrv-doi}

\bibliography{template}
\end{document}